\documentclass[conference]{IEEEtran}
\IEEEoverridecommandlockouts
\usepackage{cite}
\usepackage{amsmath,amssymb,amsfonts}
\usepackage{graphicx}
\usepackage{textcomp}
\usepackage{xcolor}
\usepackage{cite}
\usepackage{amsmath,amssymb,amsfonts}
\usepackage{graphicx}
\usepackage{textcomp}
\usepackage{xcolor}
\usepackage{float}
\usepackage{tabularx}
\usepackage{url}
\usepackage[colorlinks,urlcolor=blue,linkcolor=blue,citecolor=blue]{hyperref}

\def\BibTeX{{\rm B\kern-.05em{\sc i\kern-.025em b}\kern-.08em
    T\kern-.1667em\lower.7ex\hbox{E}\kern-.125emX}}
\begin{document}

\title{Vision Transformers and YoloV5 based Driver Drowsiness Detection Framework} 


\author{\IEEEauthorblockN{Ghanta Sai Krishna\IEEEauthorrefmark{1},
Kundrapu Supriya\IEEEauthorrefmark{2}, Jai Vardhan\IEEEauthorrefmark{3} and
Mallikharjuna Rao K \IEEEauthorrefmark{4}}
\IEEEauthorblockA{Data Science and Artificial Intelligence Department,
IIIT Naya Raipur\textbf{\IEEEauthorrefmark{1}\IEEEauthorrefmark{4}}\\
Computer science and Engineering, IIIT Naya Raipur\IEEEauthorrefmark{2}\IEEEauthorrefmark{3}\\
IEEE Member \IEEEauthorrefmark{4}\\
\IEEEauthorrefmark{1}ghanta20102@iiitnr.edu.in,
\IEEEauthorrefmark{2}kundrapu20100@iiitnr.edu.in,
\IEEEauthorrefmark{3}jai20100@iiitnr.edu.in,
\IEEEauthorrefmark{4}rao.mkrao@gmail.com}}
\maketitle

    \begin{abstract} Human drivers have distinct driving techniques, knowledge, and sentiments due to unique driving traits. Driver drowsiness has been a serious issue endangering road safety; therefore, it is essential to design an effective drowsiness detection algorithm to bypass road accidents. Miscellaneous research efforts have been approached the problem of detecting anomalous human driver behaviour to examine the frontal face of the driver and automobile dynamics via computer vision techniques. Still, the conventional methods cannot capture complicated driver behaviour features. However, with the origin of deep learning architectures, a substantial amount of research has also been executed to analyze and recognize driver's drowsiness using neural network algorithms. This paper introduces a novel framework based on vision transformers and YoloV5 architectures for driver drowsiness recognition. A custom YoloV5 pre-trained architecture is proposed for face extraction with the aim of extracting Region of Interest (ROI). Owing to the limitations of previous architectures, this paper introduces vision transformers for binary image classification which is trained and validated on a public dataset UTA-RLDD. The model had achieved 96.2\% and 97.4\% as it's training and validation accuracies respectively. For the further evaluation, proposed framework is tested on a custom dataset of 39 participants in various light circumstances and achieved 95.5\% accuracy. The conducted experimentations revealed the significant potential of our framework for practical applications in smart transportation systems.
  \end{abstract}


\section{Introduction}
Due to the lack of drowsiness detection systems in Advanced Driver Assistance Systems (ADAS), numerous drivers and pedestrians are seriously injured as result of drowsy driving.  According to the Central Road Research Institute (CRRI), fatigued drivers who fall asleep while driving are responsible for around 40 percent of all traffic fatalities and injuries. According to the report by National Highway Traffic Safety Administration (NHTSA) \cite{NHTSA}, approximately 1,00,000 accidents take place every year due to drowsy driving resulting in 2000 deaths and 70,000 injuries. And around 80\% of drowsiness-related car accidents are individual vehicle run-off-road crashes, in which a driver loses control of their vehicle and eventually leaves their lane or collides with the rear of the car ahead.\cite{8580568}. Therefore, drowsy driving is a significant and latent risk factor for road accidents. Hence, it is necessary to develop a systematic drowsiness detection algorithm in order to reduce road accidents.

Drowsy driving detection has emerged as a significant research area in recent years. According to recent studies, drowsiness detection techniques are classified into three categories: physiological measures, vehicle-based measures, and face analysis \cite{IJERT}, \cite{HASAN2021}. Firstly, physiological measures are dependent on body elements such as heart rate, body temperature, pulse rate, and so on, as physical conditions alter when a driver becomes tired \cite{article1}. In general, ECG \cite{GROMER20191938}, EEG \cite{9669234}, EMG, and EOG \cite{SONG2020101865} are the commonly used physiological signals for assessing the physical conditions of a human (driver). The main disadvantage of adopting physiological methods is that it is crucial to assure driver convenience while wearing sensors \cite{article2}. Second, vehicle-based measures detect drowsiness by observing vehicle behaviour such as steering wheel movement, random braking, speed variations, and so on. Sensors are installed in the vehicle components to monitor driving performance and identify driving trends in order to detect drowsiness. The main drawback of adopting vehicle-based methods is that the vehicle's behaviour can be altered owing to bad weather and road conditions, heavy medication, and so on \cite{s21144833}. Finally, behavioural measures or face analysis detect drowsiness by observing facial expressions and movements using machine learning and computer vision (CV) \cite{1234}.

In recent years, numerous behavioral methods have been developed to detect drowsiness. For facial expression and drowsiness detection, various researches used different algorithms. Algorithms used for detecting facial expressions include ViolaJones (Haarcascade) \cite{article3}, and canny edge detection \cite{IJERT} and neural network  algorithms like CNN \cite{9216020},  ANN \cite{Vesselenyi_2017}, Naive Bayes classifier \cite{article3}, and GAN’s \cite{9042231} can be used for drowsiness detection. In this paper, we proposed a behavioral method framework from face detection to drowsiness detection using Yolo V5 and Vision Transformers (ViT). The significant contributions of this paper are as follows: 

\begin{description}
  \item[$\bullet$] Trained customized YoloV5 pretrained model for face detection.
  \item[$\bullet$] Built and analysed the robust binary image classification model with vision transformers for drowsiness detection.
  \item[$\bullet$] Tested the framework with real-time custom dataset consisting different scenarios
\end{description} 


The following sections of this research paper are organized chronologically. Section II provides an overview of the recent drowsiness detection techniques. Section III describes the paper's approach in sequential order, beginning with face identification using YoloV5, progressing to image augmentation, and ending with image classification using Vision Transformers. Section IV provides a in detail explanation of the experimental analysis of the model and its test on custom dataset. Section V is comprehended with key findings and comparative analysis. Finally, Section VI concludes this research paper.

\section{Related Works}
Nowadays, driving is a vital and routine activity for many people, hence indepth attempts are required to understand, recognize and forecast the driving behaviour of the people. Inorder to fulfill the need, numerous experiments have taken place in order to detect unusual driver behaviour like drowsiness etc. In recent years, advanced methods have been used in order to attain improved performance of drowsiness detection systems.  Zuojin Li et al.\cite{article} examined vehicle based methods. In the method, they collected different types of drowsiness data like yaw angles and steering wheel angles by using sensors attached to vehicle The features obtained from steering wheel angles and yaw angles are examined, and approximate entropy features are calculated on time series data. To identify the drowsiness state of the driver the paper uses  approximate entropy features as input for Back-propagation Neural Networks classifier and obtained an accuracy of 87.21\%. The algorithm identifies the driver as awake, drowsy, and very drowsy. 

Behavioral methods have been introduced in recent years to overcome the problems caused by physiological and vehicle-based methods. Behavioral methods are more reliable than vehicle-based methods because they focus on the driver's facial expression rather than the vehicle's behaviour. On the other hand, Physiological methods produce highly accurate results but are not widely used due to their complicated nature.
Advancing the behavioral methods, Sherif Said et al. \cite{Said2018} proposed a drowsiness detection system which involves Viola Jones algorithm for detecting eye and face regions. The system detects the drowsiness state of the driver and sends an alarm to alert the driver. This system was tested for indoor and outdoor environments and it has shown 82\% and 72.8\% for indoor and outdoor environments respectively. Feng You \cite{8930504} constructed an algorithm that performs offline training on the driver before it can be used online. Dlib's CNN is used for detecting the face regions, and the eye aspect ratio is calculated using Dlib's 68 point facial landmarks. The algorithm used for identifying drowsiness is divided into two stages: offline training that uses SVM classifier and online monitoring for detecting the driver's state online. After a comparative analysis the accuracy of the proposed algorithm is 94.8\% but the main disadvantage is that the SVM classifier requires training from the end-users i.e, it must be trained for every driver.  In \cite{9042231}, Generative Adversarial networks (GAN) was introduced for data augmentation and Convolutional Neural Network (CNN) for prediction. After a detailed analysis, the authors concluded that usage of GAN produced new data samples (practical images) and CNN's have increased the accuracy of the model.

R Tamanani et al. \cite{9394715} proposed a method that consists of two sequential systems: the input system which employs the Haar cascade algorithm for face detection and preprocessing of real-time input video stream data, and the output system employs the CNN LeNet architecture for feature extraction and image classification. The model was evaluated using stratified 5-fold cross validation on UTA-RLDD, and it attained an average accuracy, precision, recall, and F1-score of 91.8\%, 92.8\%, and 92\%, respectively. On a custom dataset, the system has achieved an accuracy of 98\%, 84\%, and 88\% on the training, validation, and testing data, respectively.

In our proposed framework we explored novel architectures YoloV5 \cite{7748350}, \cite{qi2022yolo5face} and vision transformers \cite{DBLP:journals/corr/abs-2010-11929}, \cite{9658539}, \cite{9531646}, \cite{rs13030516} which have advantages over existing frameworks. To analyze the performance in real-time, the proposed framework is tested with custom dataset.

\section{Research Methodology}
\begin{figure}[ht]
\centerline{\includegraphics[height=70mm,width=85mm]{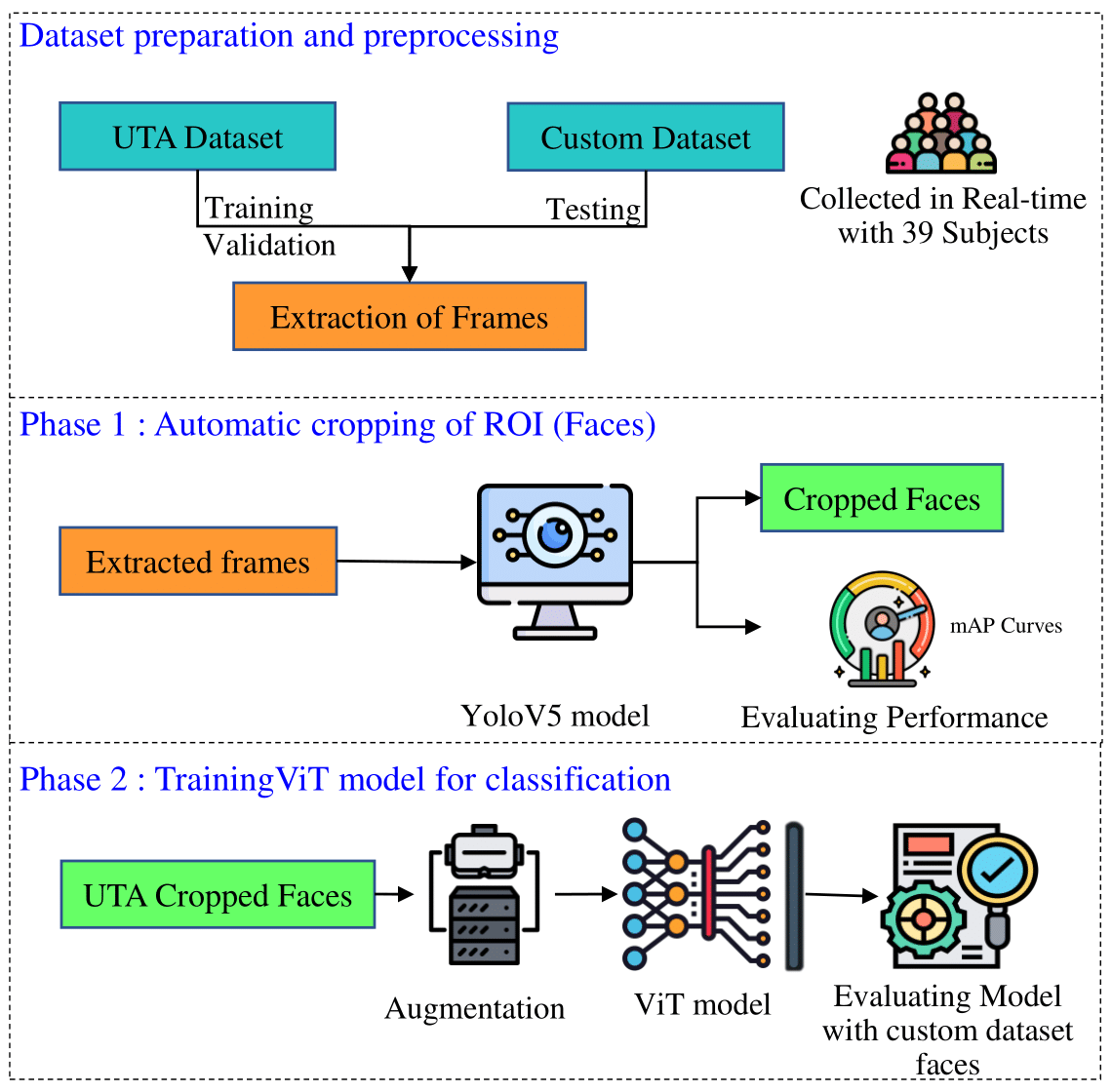}}
\caption{Overview of proposed framework}
\label{fig1}
\end{figure}

 This section discusses our proposed novel framework, which mainly consists of two phases. A custom YoloV5 pre-trained model and self-trained ViT architecture are proposed for automated face extraction, and binary image classification. We also review data augmentation which is performed to increase the data for obtaining a skillful ViT model. The illustration of a concise schema of the framework consisting drowsiness recognition algorithms is in Fig. \ref{fig1}.

The proposed system utilizes two video datasets: 1) University of Texas at Arlington Real-Life Drowsiness Dataset (UTA-RLDD) 2) Custom dataset collected. The proposed system works on a single frame in detecting drowsiness. Hence, frames are extracted from the video dataset and serve as the framework's input. 

\subsection{Face detection with YoloV5}\label{A2}
As most of the extracted frames from both datasets are wide-angle frames, there is a need to extract Region of Interest (ROI) from the wide-angle frames to improve the model's performance. For face detection and computerised cropping of faces from wide-angle frames, the YoloV5 framework is utilized. The YoloV5 architecture is the amalgamation of Cross Stage Partial network (CSP) and the Darknet. The weights and the configurations for YoloV5 architecture are accomplished by annotating ROI for set of wide-angle frames in the UTA-RLDD dataset. The CSP and Darknet grid does the feature extraction and target information extraction from the annotated input image. During face identification, the input vector is divided into $A\times A$ grids. If the centre of the target is in a grid, then that particular grid is accountable for the target identification. The location of the regression box of the face can be obtained by following \eqref{eq1} :

\begin{equation}
    C_{i}^{j} = P_{i, j} \times IOU_{Predicted}^{True}  \label{eq1}
\end{equation}

Where, $C_{i}^{j}$ symbolises the confidence score of the $j$th bounding box of the $i$th grid. $P_{i, j}$ denotes whether there exists a target in the $j$th bounding box of the $i$th grid. If the target is within the bounding box, then $P_{i, j}$ will be equal to one if not zero. The $IOU_{Predicted}^{True}$ is widely utilised parameter that characterises the Intersection Over Union (IOU) between the true box and the estimated box. The more precise the location of the predicted box is predicted when the IOU score is high.

\subsection{Image Augmentation}
\label{A3}
\begin{figure}[!ht]
\centerline{\includegraphics[height= 55 mm,width=70mm]{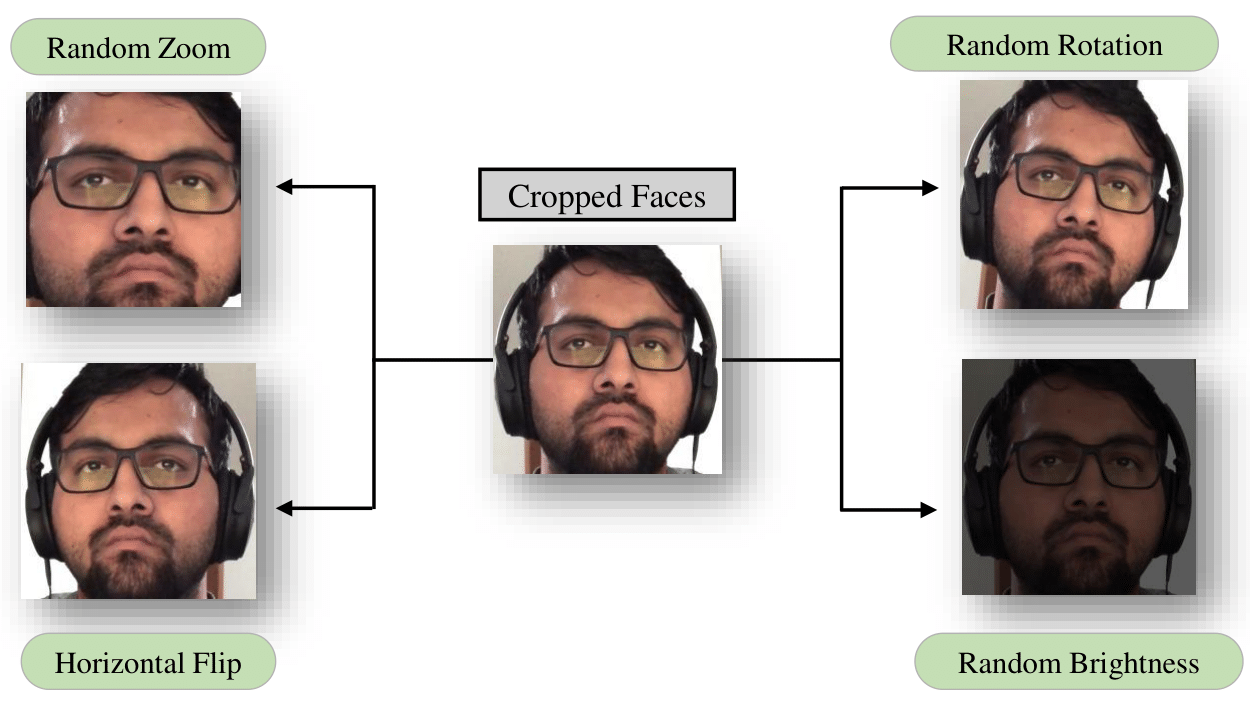} }
\caption{Augmentation of a sample cropped face}
\label{fig2}
\end{figure}

Training Vision Transformer models on more image data can result in better accurate and skilful models. The Image augmentation techniques can generate variations of the images that can enhance the mastery and performance of the ViT frameworks. Image augmentation is a technique to artificially create new training images from existing training images. The image transformations include various operations like shift, flip, zoom, etc. The Fig. \ref{fig2} represents the augmentation of a sample cropped face. The Table \ref{table1} represents the configuration for the data augmentation to generate new training image data.

\begin{table}[ht]
\caption{Data augmentation factors used}
\label{table1}
\begin{center}
\scalebox{1}
{
\begin{tabular}{|c|c|}
\hline
\textbf{Technique} & \textbf{Hyper-parameter}  \\ 
\hline
Normalization & $\textcolor{green}{\checkmark}$\\
\hline
Resize &  $\textcolor{green}{\checkmark}$\\
\hline
Random Flip & Horizantal\\
\hline
Random Rotation & Factor = 0.01\\
\hline
Random Zoom & Factor = 0.2\\
\hline
Random Brightness & Range = [0.2, 1.0]\\
\hline
\end{tabular}
}
\end{center}
\end{table}

\subsection{Vision Transformers for Image Classification}\label{A1}

After executing face detection and image augmentation, there is a need for efficient machine learning architecture to fulfil image classification. The current research methodology uses vision transformers for effective image classification. The ViT framework utilizes the transformer model with successions of image speckles without involving the convolution layers in the framework. Perspective to the traditional transformers, an input frame is resized and converted into N patches and delivers the successions of linear embeddings of these separated patches as an input to the ViT architecture. 

\begin{equation}
    I \in \mathbb{R}^{H\times W\times C}  \Rightarrow I_{P} \in \mathbb{R}^{N\times (P^{2}.C)} \label{eq2}
\end{equation}

\begin{equation}
    N = H\times W / P^{2}
    \label{eq3}
\end{equation}

\begin{figure}[ht]
\centerline{\includegraphics[height=39mm,width=90mm]{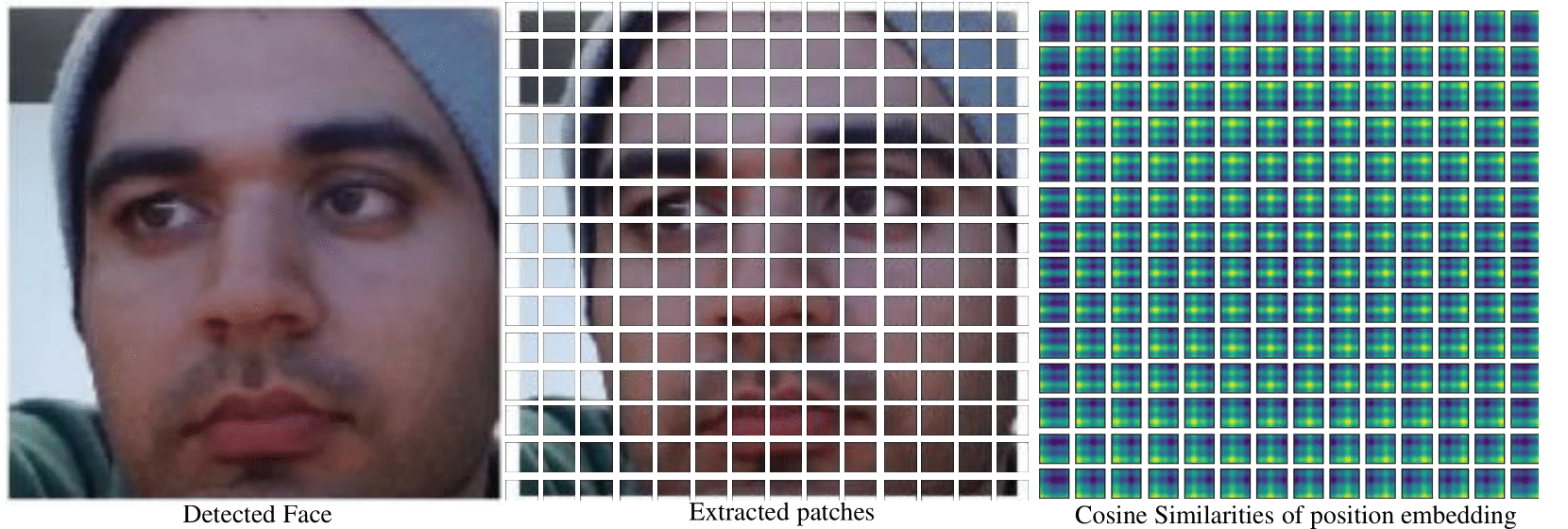} }
\caption{Visualization of the patches and embeddings}
\label{fig3}
\end{figure}

In the above equations \eqref{eq2} \eqref{eq3}, $H, W, C$ represents the height, width and channels of the original image, $(P, P)$ is the resolution of a every patch and $N$ represents total number of resulting patches. The illustration of the speckles created from the sample image is shown in Fig. \ref{fig3}. The characteristics of the patches created from the image are in the Table \ref{table2}.

\begin{table}[!ht]
\begin{center}
\caption{characteristics of the created patches}
\label{table2}

\scalebox{1}
{
\begin{tabular}{|c|c|}
\hline
\textbf{Technique} & \textbf{Skewness}  \\ 
\hline
Size of image & 392 $\times$ 392\\
\hline
Size of patch & 28 $\times$ 28\\
\hline
Patches per image & 196\\
\hline
Elements per patch & 2352\\
\hline
\end{tabular}
}
\end{center}
\end{table}

\begin{figure*}[ht]
\centerline{\includegraphics[height=98mm,width=175mm]{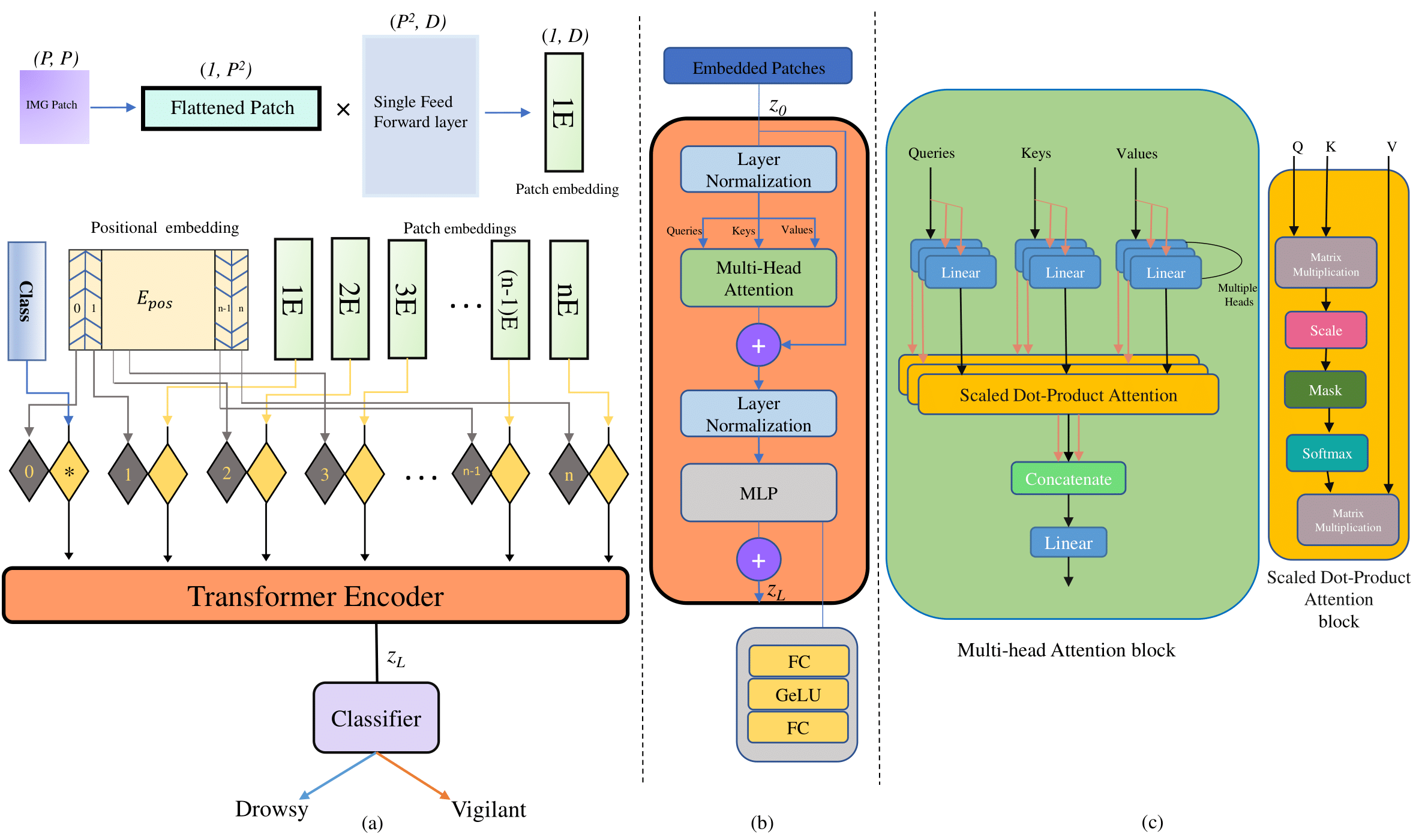}}
\caption{(a) Generation of patch embeddings and Conceptual overview of ViT model (b) Transformer Encoder (c) Multi head Attention block}
\label{fig4}
\end{figure*}

Furthermore, the matrices of every patch with dimensions ($P, P$) are flattened into matrices of dimensions ($1, P^{2}$). To obtain a linear patch projection, these flattened patches ($E_{(1, P^{2})}$) are transmitted via a single feed forward layer which contain embedding matrix ($F_{(P^{2}, D)}$). After projecting, the patches transformed into a constant latent vector of size D (projection dimension) which are called as patch embeddings ($E_{(1, D)}$).  The generation of patch embeddings and conceptual overview of ViT model is visualized in Fig. \ref{fig4}(a). A learnable [\textit{class}] embedding is prehended to linear patch projections to assist the classification head. As the data is sent at an instance, the order of successions is not implemented naturally. To resolve this problem, an erratically forged positional embedding matrix ($E_{pos}$) is added to the concatenated matrix consisting of the patch embeddings ($E$) and the learnable class embedding ($I_{class}$). The resultant embedded successions of extracted patches with the token $z_{0}$ is delivered in \eqref{eq4a}\eqref{eq4}.

\begin{equation}
 \textbf{z}_{0} = [I_{class}; x_{1}\textbf{E}; x_{2}\textbf{E}; . . . ; x_{n}\textbf{E}] + \textbf{E}_{pos}         
 \label{eq4a}
\end{equation}
where
\begin{equation}
 \textbf{E} \in \mathbb{R}^{(P^{2}C)\times D},\textbf{E}_{pos} \in \mathbb{R}^{(N+1)\times D}
 \label{eq4}
\end{equation}

These embedding sequences together are input for transformer encoder, which consists of $L$ identical layers of Multi-headed Self-Attention (MSA) blocks and Multi-Layer Perceptron (MLP) blocks as shown in Fig. \ref{fig4}(b). Both subcomponents of the transformer encoder operates with normalization layer (LN) succeeded by residual skip connections as shown in mathematical expressions \eqref{eq5}, \eqref{eq6}.

\begin{equation}
\textbf{z}^{`}_{l} = MSA(LN(z_{l-1})) + z_{l-1}
    \label{eq5}
\end{equation}

\begin{equation}
\textbf{z}_{l} = MLP(LN(z^{`}_{l})) + z^{`}_{l}
    \label{eq6}
\end{equation}

The procedures involved in MSA block are represented in Fig. \ref{fig4}(c). In the attention block, the input vectors are stacked and multiplied by an initial batch of weights $W_{q}, W_{k},$ and $W_{v}$ to produce three individual matrices of queries Q, keys K, and values V. A dot product is calculated from every query q from Q and all the keys k in K, i.e., the matrix Q is multiplied with the transpose of the matrix K to compute the attention matrix. The scaling dot-product executed in the Self Attention (SA) block is equivalent to the regular dot-product, but it integrates the dimension of the key $d_{k}$ as a scaling factor. The attention weight is calculated by feeding the dot product results to the softmax. The scaled dot product attention for a head is estimated by multiplying attention weight with the value of each patch embedding's vector as represented in \eqref{eq7}. The MSA block in the utilised transformer encoder calculates the scaled dot-product attention individually for h heads. The outcomes of all of the attention heads are concatenated together and then passed through a feed-forward layer with learnable weights $W^{0}$ as shown in \eqref{eq8}.

\begin{equation}
SA = Softmax(\frac{QK^{T}}{\sqrt{d_{k}}})\times V = W_{attention} \times V
    \label{eq7}
\end{equation}

\begin{equation}
MSA = concat(SA_{1}; SA_{2}; ..... SA_{h})\times W^{0}
    \label{eq8}
\end{equation}
\[W^{0} \in \mathbb{R}^{hd_{k} \times D}\]

The MLP block consists of fully connected feed-forward dense layers with GeLU non-linearity. At final layer of encoder, the foremost element in the sequence $z_{L}$  is passed to an external lead classifier for estimating the class labels.

\section{Experimental Analysis and results}

\subsection{Dataset Preparation and Utilization}
This experiment utilizes a drowsiness estimation benchmark dataset UTA-RLDD \cite{ghoddoosian2019realistic}, which is one of the most considerable dataset among available public datasets, for training the ViT model. The training UTA-RLDD dataset includes 36 participants with distinct scenarios. The videos in dataset correspond to the two most essential situations: drowsiness-related symptoms (yawning, nodding) and a combination of non-drowsiness related activity (talking, laughing, gazing at both sides), individually lasting around one and half minutes. The random frames of each participant are collected and binarily labelled based on their drowsiness status as either active or fatigued. The resolution of the frames in this dataset is $640\times 480$, which is relatively high compared to other drowsiness detection datasets. The faces in the dataset have significant transformations in scale, pose, and perceptions. Thus, this benchmark dataset is appropriate to demonstrate the efficiency and performance in practical scenarios. This image dataset is operated for the annotating and training the YoloV5 and ViT models. 

The testing of the complete framework is entirely performed by custom dataset. The dataset is collected for 39 subjects with high resolute DSLR camera. The variations in posture, angle of view, and orientation of capturing of the subjects are increased in comparision with UTA-RLDD Dataset. The specifications of our custom dataset and UTA-RLDD dataset is given below in Table \ref{table3}

\begin{table}[ht]

\caption{Comparision of the datasets}
\label{table3}
\begin{center}
\scalebox{1}
{
\begin{tabular}{|c|c|c|}
\hline
\textbf{Attribute} & \textbf{UTA-RLDD Dataset} & \textbf{Our Dataset}  \\ 
\hline
Frame resolution & $640\times 480$ & $3840 \times 2160$\\
\hline
Number of Subjects & 36 & 39\\
\hline
Collected in day and night & $\textcolor{red}{\times}$ & $\textcolor{green}{\checkmark}$ \\
\hline
Multi-oriental frames & $\textcolor{red}{\times}$ & $\textcolor{green}{\checkmark}$ \\
\hline
Number of scenarios & Five & Nine\\
\hline
Number of frames & 9180 & 1246\\
\hline
Utilization & Training, Validation & Testing\\
\hline
\end{tabular}
}
\label{tab1}
\end{center}
\end{table}

\subsection{Computation Specifications of proposed system}
This section deals with both the hardware and software
computational specifications of our driver drowsiness detection framework. Tensorflow 2.0, Keras, and OpenCV libraries were utilised to build the framework, which was written in Python 3.9. The YoloV5 and ViT models are not trained using a high-performance graphics processing unit. The specifications  and minimum requirements for training and testing the framework are depicted in Table \ref{table4}. 

\begin{table}[ht]
\caption{}
\label{table4}
\begin{center}
\scalebox{1}
{
\begin{tabular}{|c|c|}
\hline
\textbf{Specifications} & \textbf{System's Configuration}  \\ 
\hline
Operating system & Ubuntu 20.04.3 LTS\\
\hline
CPU & Intel® i7 10th gen\\
\hline
RAM & 15.8 Usable\\
\hline
GPU & Intel® UHD Graphics\\
\hline
Frameworks & Tensorflow, OpenCV\\
\hline
\end{tabular}
}
\end{center}
\end{table}

\subsection{Evaluating YoloV5 for face detection}
\begin{figure}[ht]
\centerline{\includegraphics[height=57mm,width=84mm]{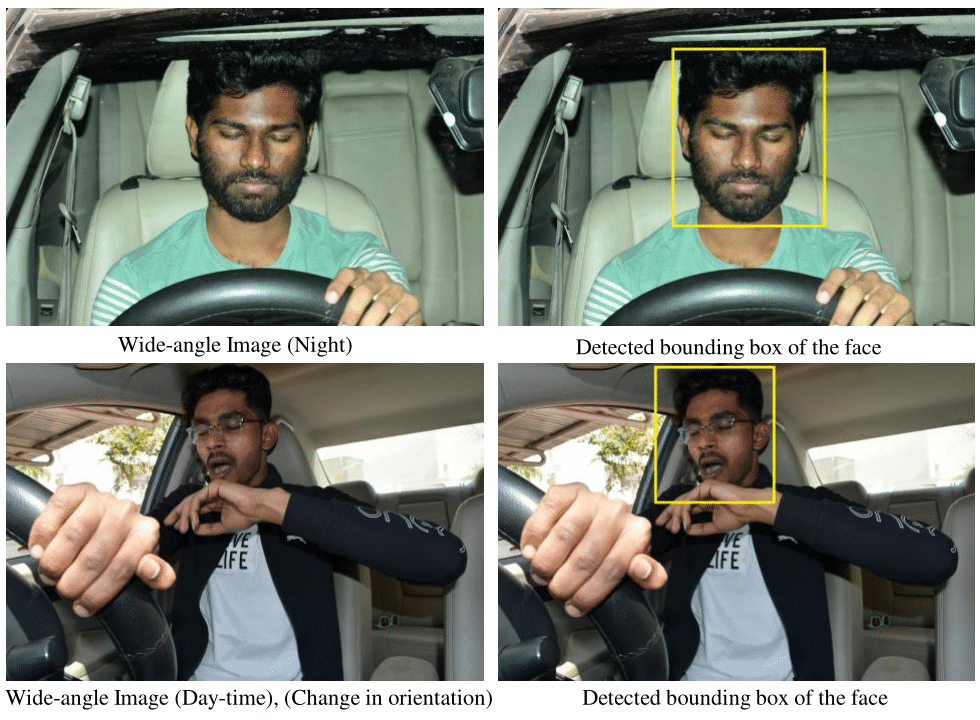} }
\caption{YoloV5 results on sample images in custom dataset}
\label{fig5}
\end{figure}

\begin{figure}[ht]
\centerline{\includegraphics[height= 45 mm,width=80mm]{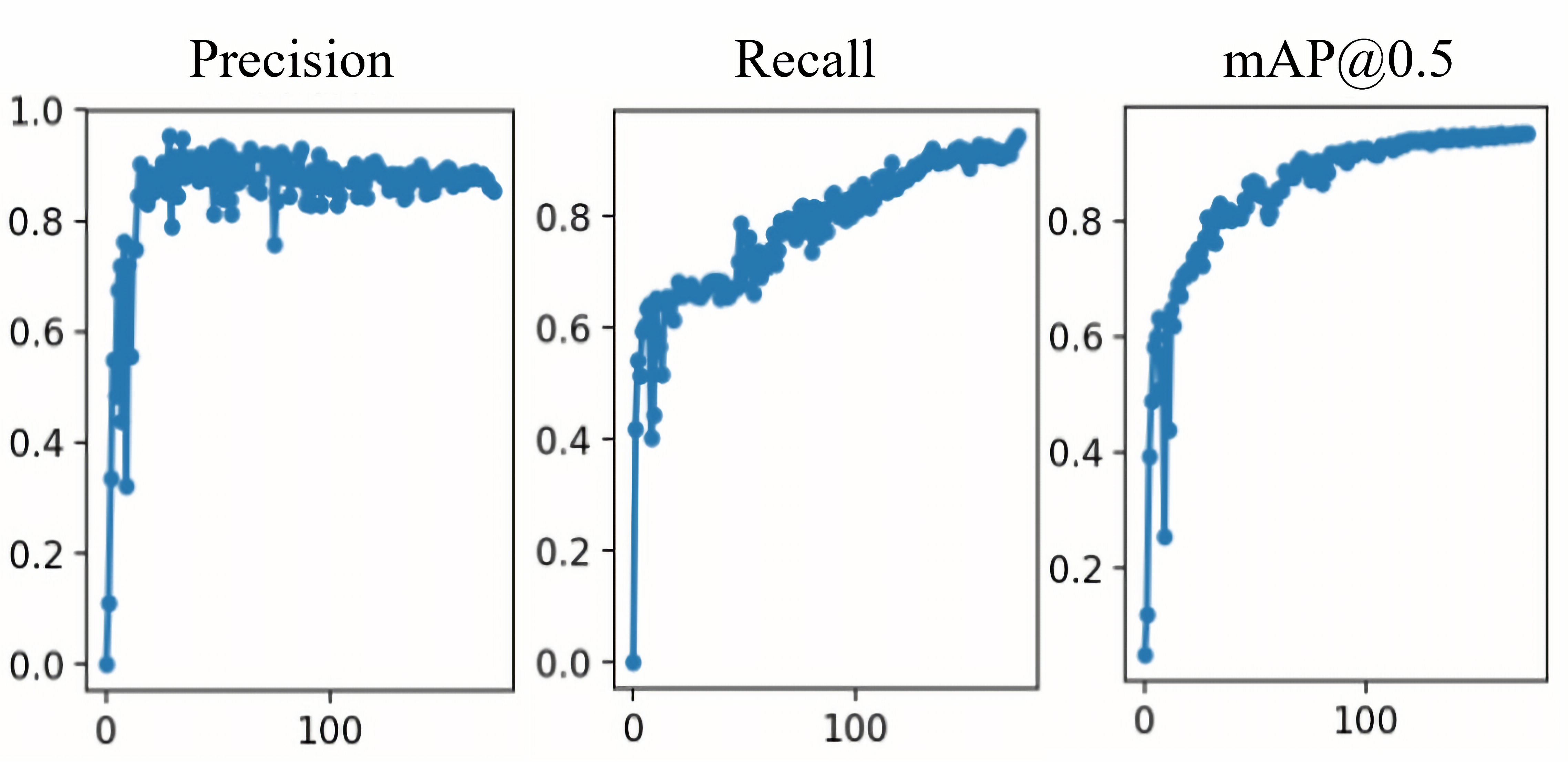} }
\caption{Visualization of YoloV5 Performance for 200 epochs}
\label{fig6}
\end{figure}

The YoloV5 is implemented as a core architecture with all the configurations for accurate face detection and to extract ROI from the input image. We had trained the model with customized settings with 200 epochs. After several tests, the minimum confidence score for the model is set to be 0.75. The precision, recall and mean average precision metric(mAP) at 0.5 threshold curves represented in Fig. \ref{fig6}. The detection effect of the trained architecture on a samples of the dataset is shown in Fig. \ref{fig5}. After numerous tests, the speed of the trained YoloV5 has been calculated as 51.9  images per second. After obtaining the location of the bounding boxes, the ROI is cropped from the wide-angle frames of both training and validation frames.

\begin{figure*}[ht]
\centerline{\includegraphics[height=50mm,width=185mm]{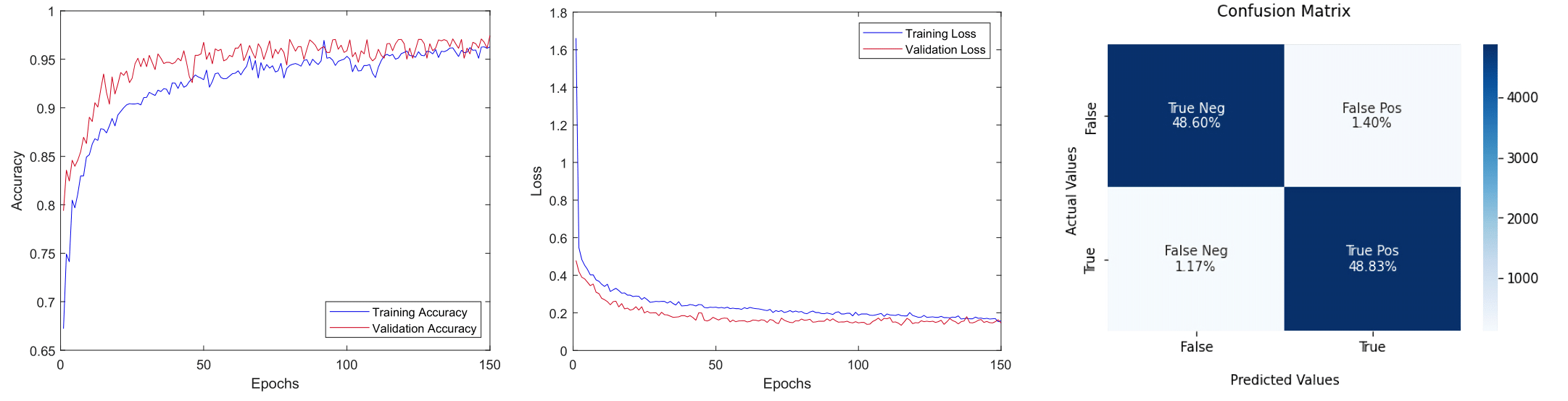}}
\caption{Learning curves of training and validation accuracy and loss for 150 epochs and its confusion matrix}
\label{fig7}
\end{figure*}

\subsection{Evaluating trained ViT architecture}

After face detection, the training set is augmented with efficient methods, as discussed in Section III. In this experiment, the ViT model is trained for drowsiness state classification is binary as drowsy or alert.

The typical benchmarks for evaluating classification models are used in evaluating our ViT framework. The learning curves of accuracy and loss for training and validation of the models are represented in Fig. \ref{fig7}. These learning plots demonstrate a good fit learning algorithm as the validation and training curves both maintain a point of stability with minimal gap. In order to maximize performance, the training of the efficient ViT model was incorporated with three missions simultaneously: 1) computation of output, 2) debugging the errors, and 3) tuning the hyper-parameters. After multiple iterations of tuning the hyper-parameters, the maximum training and validation accuracies are 96.2\% and 97.4\% respectively which are achieved with a specific set of hyper-parameters, as shown in the Table \ref{table5}.

\begin{table}[ht]
\caption{Tuned Hyperparameters of ViT model}
\begin{center}
\label{table5}
\scalebox{1}
{
\begin{tabular}{|c|c|}
\hline
\textbf{Hyper-parameter} & \textbf{Attribute}  \\ 
\hline
Number of classes & 2\\
\hline
Input shape & (256, 256, 3)\\
\hline
Resized image Size & (392, 392)\\
\hline
Patch Size & 28\\
\hline
Batch Size & 256\\
\hline
Number of Epochs & 150\\
\hline
Learning Rate & 0.001\\
\hline
Weight Decay & 0.0001\\
\hline
Number of Heads & 4\\
\hline
Transformer Layers & 8\\
\hline
Transformer Units & [128, 64]\\
\hline
MLP head Units & [2048, 1024]\\
\hline
\end{tabular}
}
\end{center}
\end{table}

For further evaluation of the performance of the classification ViT model, hamming loss and Binary Cross-Entropy are calculated. The calculated cross-entropy loss and the hamming loss for the trained ViT model are $0.6907$ and $0.0673$ respectively. Being closer to zero, the log loss had indicated good results. Cross entropy loss had penalized inaccurate predictions excessively, which is a relevant factor for a loss function, but undesirable for a metric. Hence, accuracy metrics like Precision, Recall, and F1 scores were estimated and represented. The precision, sensitivity (recall), F1-score are estimated based on the following eq \eqref{eq9}, \eqref{eq10},\eqref{eq11} respectively.

\begin{equation}
Precision = \frac{T_{P}}{T_{P} + F_{P}}
    \label{eq9}
\end{equation}

\begin{equation}
Sensitivity = \frac{T_{P}}{T_{P}+F_{N}}
    \label{eq10}
\end{equation}

\begin{equation}
F1Score = 2 \times \frac{Precision \times Recall}{Precision + Recall}
    \label{eq11}
\end{equation}

where $T_{P}$ is the number of images that are accurately
determined as the drowsiness state; $F_{N}$ is
the number of images that inaccurately recognized
as the non-drowsiness state; $F_{p}$ is the
number of non-drowsiness images that erroneously determined
as the drowsiness state; $T_{N}$ is the number
of non-drowsiness images that precisely recognized as the non-
drowsiness state. The calculated values of precision, recall, F1-score are shown in the Table \ref{table6}.

\begin{table}[ht]
\caption{Calculated Evaluation metrics}
\label{table6}
\begin{center}
\scalebox{0.9}
{
\begin{tabular}{ccccc}
\hline
 & Precision & Recall & F1 Score & Image Support \\ 
\hline
Drowsy & 0.97 & 0.98 & 0.97 & 626\\
\hline
Vigilant & 0.98 & 0.97 & 0.97 & 620\\
\hline
Macro Average & 0.97 & 0.97 & 0.97 &  1246\\
\hline
Weighted Average & 0.97 & 0.97 & 0.97 &  1246\\
\hline
\end{tabular}
}
\end{center}
\end{table}

\subsection{The Influence of Training and Validation Data Splits on Test Accuracy}

The primary objective is to determine how the performance of ViT model predictions on our custom dataset changes with various combinations of training and validation data sets. In order to experiment, we selected four combinations of data splitting for the UTA dataset. We validated and trained the ViT model on each set (80-20, 70-30, 60-40 and 50-50) and compared the model's accuracy after validation and training. 

\begingroup
\setlength{\tabcolsep}{5pt}
\begin{table*}[t]
	\centering
	\caption{Statistical Information of performance of ViT model with custom Dataset}
	\label{table7}
	\begin{tabular}{|c|c|c|c|c|c|c|c|c|c|}
		\hline
		
		\multicolumn{1}{|c}{\textbf{}} &
		\multicolumn{3}{|c|}{\textbf{Day-Time}}  & \multicolumn{3}{c|}{\textbf{Evening-Time}}  & \multicolumn{3}{c|}{\textbf{Night-Time}}     \\
		
		\hline
		\centering Scenario &
		\centering Drowsy (F)&
		\centering Vigilant (F)& 
		\centering Accuracy & 
		\centering Drowsy (F)&
		\centering Vigilant (F)& 
		\centering Accuracy & 
		\centering Drowsy (F)&
		\centering Vigilant (F)& 
		\centering Accuracy(F) \tabularnewline
		\hline

		\centering Bare Face & 
		\centering 0.979 & 
		\centering 0.978 & 
		\centering 0.979 & 
		\centering 0.985 & 
		\centering 0.979 & 
		\centering 0.982  & 
		\centering 0.949 & 
		\centering 0.954 & 
		\centering 0.951 \tabularnewline
		\hline
		
		\centering Spectacles & 
		\centering 0.955 & 
		\centering 0.959 & 
		\centering 0.957 & 
		\centering 0.989 & 
		\centering 0.979 & 
		\centering 0.984  & 
		\centering 0.892 & 
		\centering 0.917 & 
		\centering 0.905 \tabularnewline
		\hline
		
		\centering Sunglasses & 
		\centering 0.932 & 
		\centering 0.955 & 
		\centering 0.943 & 
		\centering 0.965& 
		\centering 0.970 & 
		\centering 0.967 & 
		\centering - & 
		\centering - & 
		\centering - \tabularnewline
		\hline
		
		\centering Average & 
		\centering 0.955 & 
		\centering 0.964 & 
		\centering 0.959 & 
		\centering 0.980 & 
		\centering 0.976 & 
		\centering 0.978  & 
		\centering 0.921 & 
		\centering 0.935 & 
		\centering 0.928 \tabularnewline
		\hline

	\end{tabular}
\end{table*}

\begingroup
\setlength{\tabcolsep}{5pt}
\begin{table*}[t]
	\centering
	\caption{Comparision of our proposed framework with the existing models}
	\label{table8}
	\begin{tabular}{|c|c|c|c|c|c|c|c|c|c|}
	\hline
Research by & Dataset Used & Facial Detector & Classifier & Overall Accuracy & Testing in real time\\ 
\hline

Bakheet et al. \cite{article3} & NTHU & Haar Cascades & HOG, Naive Bayesian & 85.62 \% &  $\textcolor{red}{\times}$\\
\hline

Shreyans M et al.\cite{9596358} & UTA & Dlib & Logistic Regression & 75.67 \% & $\textcolor{red}{\times}$\\

\hline
R Tamanani et al. \cite{9394715} & UTA & Haar Cascades & LeNet CNN & 91.8 \% & $\textcolor{green}{\checkmark}$\\ 

\hline
Anh-Cang et al. \cite{app11188441} & Mixed & SSD Network & MobileNet-V2 and ResNet-50V2 & 97\% &  $\textcolor{red}{\times}$\\
\hline
\textbf{Proposed Framework} & \textbf{UTA, Custom dataset} & \textbf{YoloV5} & \textbf{Vision Transformers} & \textbf{97.4 \%} &  $\textcolor{green}{\checkmark}$\\
\hline
	\end{tabular}
\end{table*}

\begin{figure}[h]
\centerline{\includegraphics[height=50mm,width=75mm]{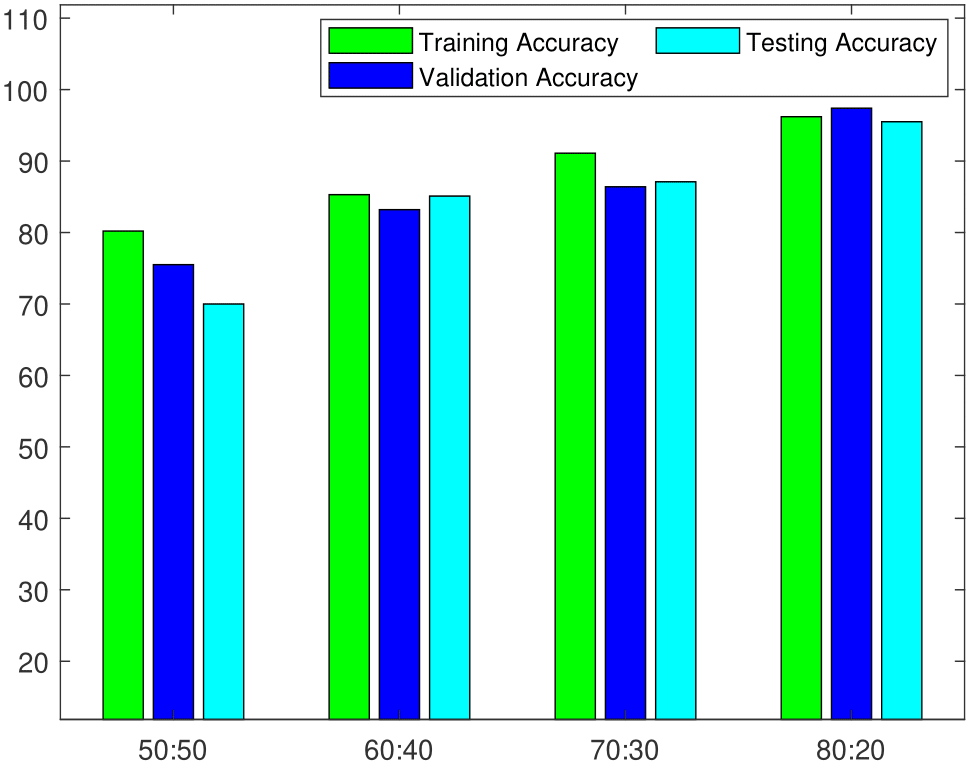} }
\caption{The Impacts of Data Splits}
\label{fig8}
\end{figure}

Furthermore, we noticed significant differences in accuracy for the train and validation split ratios, which significantly influenced the testing accuracy on the custom dataset. Finally, the model's best training, validation, and testing accuracies are 96.2, 97.4 and 95.5, respectively, with a split of 80-20, as seen in Fig. \ref{fig8}

\subsection{Evaluation of ViT model with custom dataset}
The custom dataset we had made consists of three scenarios bare face, spectacles and sunglasses with various gestures collected at day time, evening time, and night time. The proposed system is tested with the custom dataset for evaluation. Statistical analysis is performed for the model's performance on all these scenarios individually, as shown in Table \ref{table7}. There are substantial differences in the average accuracies of the model for the daytime, evening, and nighttime images in the custom dataset. The average accuracies of daytime, evening, and nighttime images are 95.9\%, 97.8\% and 92.8\%. The model had performed better in evening time than daytime and nighttime due to optimum light conditions. The final ViT model accuracy for the custom dataset is 95.5\%.

\subsection{Comparison with existing models}
An appropriate combination of architectures for face detection and classification is important to attain optimal performance. To achieve ideal performance, many frameworks are presented with distinct computer vision and machine learning architectures.
The primary focus of this subsection is the examination of the most delinquent research attempts in identifying human drowsiness using machine learning architectures. The comprehensive comparison analysis of classification techniques for human drowsiness are presented in Table \ref{table8}. Many face detection algorithms, ranging from haar cascades to CNNs, have been developed in recent drowsiness detection research endeavours. These endeavours had also   used a variety of image classification algorithms, ranging from Bayesian classifiers to CNNs.

\section{Author Contributions: Key Findings and Comparative Analysis}
This section discusses the significant findings as well as a comparative examination of our proposed model with other existing research in this field. In this section, Our proposed framework is compared against two distinct factors: the dataset and the methodology.

\textbf{Dataset:} The capability of driver drowsiness detection systems is heavily reliant on a several factors in the dataset to be analysed. A massive quantity of data is involved, indicating the greater difficulty in evaluating the system in real-time \cite{ghoddoosian2019realistic}\cite{123456}. This is also due to the ambiguity of the kind of image data in public databases. To address the aforementioned issues, we gathered a customised dataset based on real-time requirements, which resulted in optimal testing results. This dataset is also simple to analyze, and there is a good chance that it would be expanded as many investigators on drowsiness detection systems around the globe may contribute considerably.

\textbf{YoloV5 and Vision Transformers:} Most of the recent research for drowsiness detection is based on Convolutional neural networks (CNN) \cite{9216020}, Generative Adversarial Networks (GAN) \cite{9042231}, and traditional computer vision techniques\cite{1234}. We had proposed a novel framework of YoloV5 and Vision transformers for drowsiness detection in real-time which is never implemented before. Various experimentations are performed with the real-time custom dataset to identify the efficiency of the ViT architecture. In this paper, we present several extensive comparisons between our proposed system and existing solutions, which are based on the model's performance.

\section{Conclusion}
In this paper, we have introduced vision transformers for efficient driver drowsiness state estimation. This novel framework mainly comprises two sequential components; the initial component executes automatic face cropping using YoloV5 pre-trained face detection CNN architecture, and the conclusive component performs binary image classification with vision transformer (ViT) architecture. After several comprehensive analyses, the YoloV5 achieved around 95\% mAP score. The evaluation of the ViT model indicated that the framework had attained high values of average precision, sensitivity, and F1-score, which are 0.97, 0.98, and 0.97, respectively. In addition, employing a custom dataset indicates that the ViT architecture had accomplished 95.5\% accuracy on testing.

The limitation of the proposed architecture can be outlined as follows. Firstly, though the proposed model accomplished satisfactory detection performance, it requires high amount of data with labelled scence conditions for model to be trained.

In future works, several arguments can be taken into account. First, we will optimize the network configuration in the proposed architecture for use in micro-computing systems to decrease the economic expense and improve the computational efficiency without deterioration. Second, we will use generative adversarial networks to enlarge the size of training data to increase the model's performance.

\bibliographystyle{IEEEtran}
\bibliography{Main.bib}

\begin{IEEEbiography}[{\includegraphics[width=1in,height=1.25in,clip,keepaspectratio]{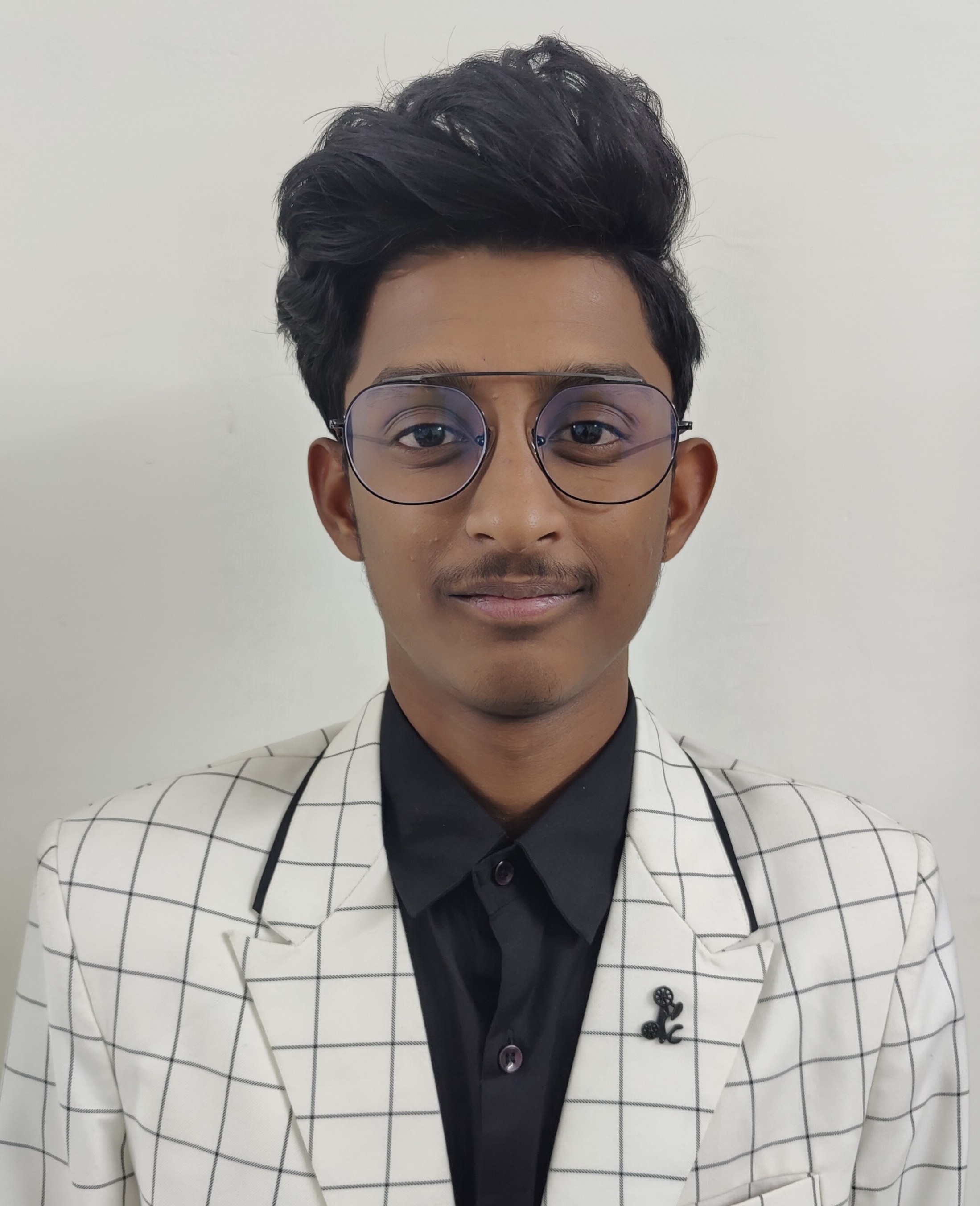}}]
{Ghanta Sai Krishna } currently pursuing bachelor's degree in Data Science and Artificial Intelligence at International Institute of Information Technology Naya Raipur. He is a passionate machine learning researcher who is more inclined towards implementing end-to-end deep learning architectures, particularly based on computer vision. He has proficient command on various deep learning architectures, training artefacts, ML model optimization, CV, NLP, statistical modelling and ML mathematical methods. He had completed various online certification courses like data analysis in python, machine learning in python offered by IBM. He is currently working on interpreting and building the research projects in the domain of modern deep learning for real-time and software defined networking with ML.
\end{IEEEbiography}

\begin{IEEEbiography}[{\includegraphics[width=1in,height=1.25in,clip,keepaspectratio]{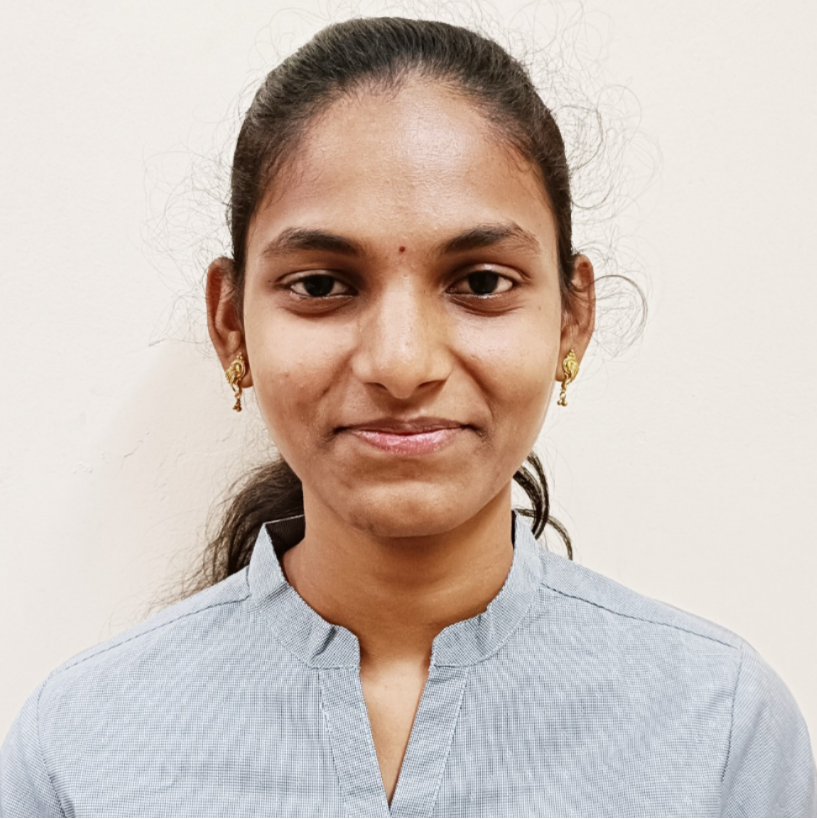}}]
{Kundrapu Supriya} 
currently pursuing Bachelor of Technology in Computer Science and Engineering at
International Institute of Information Technology, Naya Raipur, Chhattisgarh, India. She is passionate about human-centric machine learning architectures and their impact on business and research transformation. She had specialized in deep learning, and computer vision libraries such as scikit-learn, TensorFlow, Pytorch, OpenCV and OpenAI gym. Working in a fast-paced environment, rapid ideation, and development of AI-based software are a few of her distinguishing characteristics. She is keenly interested in research projects based on modern machine learning in computer vision, natural language processing and network security.
\end{IEEEbiography}

\begin{IEEEbiography}[{\includegraphics[width=1in,height=1.25in,clip,keepaspectratio]{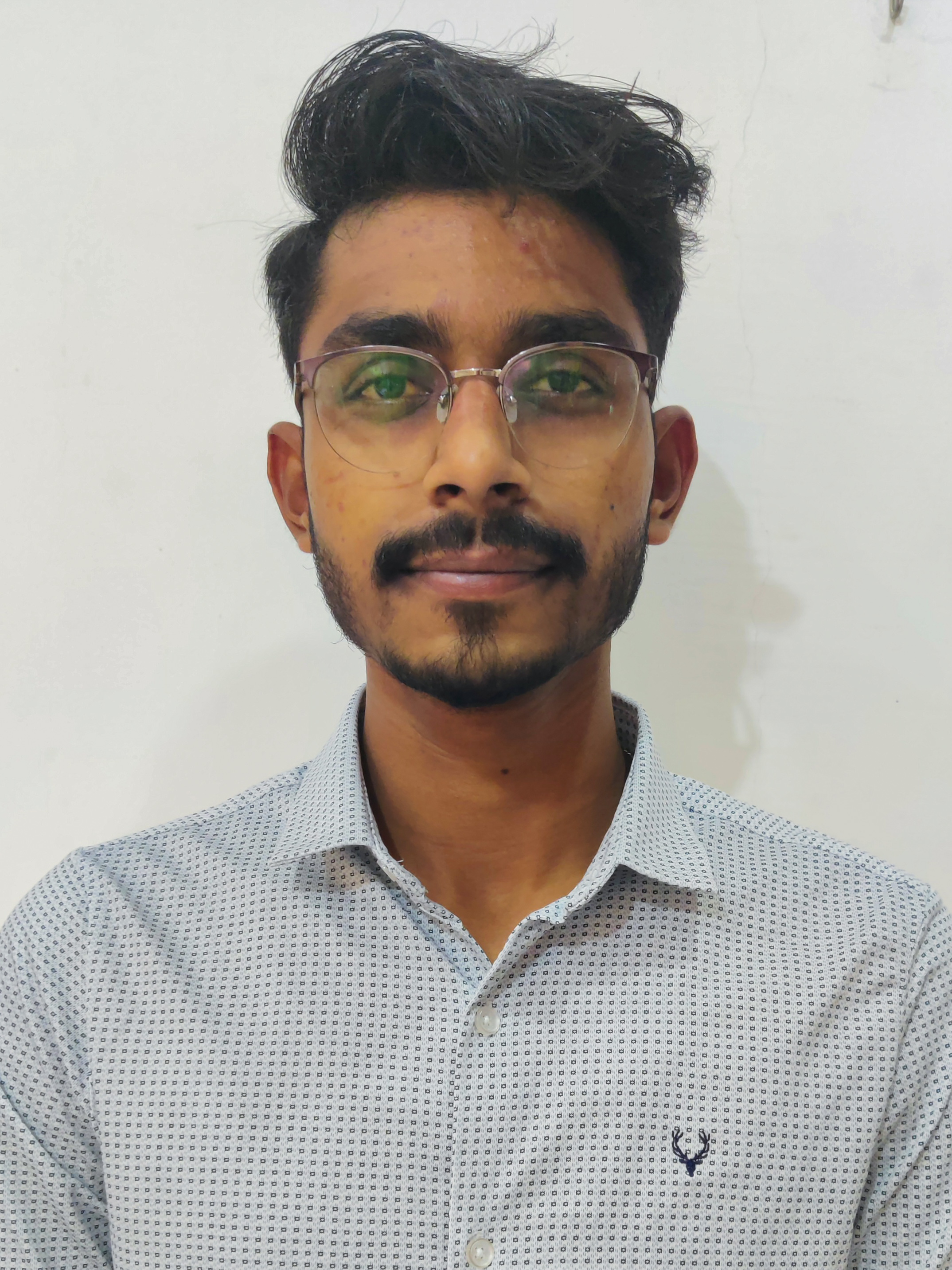}}]
{Jai Vardhan} 
'18 grew up in Hyderabad and is currently enrolled at IIIT Naya Raipur in the second year of his Bachelor's degree programme in Computer Science and Engineering. In 11th grade, Jai was the school's vice captain, and he deserved to be a leader. He's a key member of Shutterbug, a photography group at IIIT Naya Raipur that has helped capture a number of high-quality bespoke datasets. He earned a badge in the Learn to Earn Cloud Security Challenge after completing a python course that culminated in a real-time project. He's now working on a project that uses Vision Transformers to detect damaged leaves.
\end{IEEEbiography}

\begin{IEEEbiography}[{\includegraphics[width=1in,height=1.25in,clip,keepaspectratio]{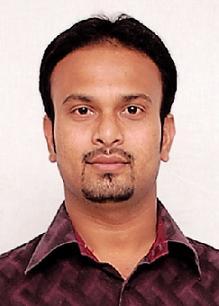}}]
{Mallikharjuna Rao K} 
currently working as Assistant Professor in the department of Data Science and Artificial
Intelligence at International Institute of Information Technology Naya Raipur (IIIT Naya Raipur), Chhattisgarh, India. He obtained bachelor’s degree in Information Science and Technology Engineering from Acharya Nagarjuna University, Guntur, Andhra Pradesh, India in 2004. Then obtained master’s degree in Computer Science from Jawaharlal Nehru Technological University Hyderabad, India in 2010, and obtained Ph.D. in Computer Science and Engineering from Jawaharlal Nehru Technological University Hyderabad, India in 2018. He received best researcher award from VIT-AP University for 2019 \& amp; 2020. He published 2 Indian patents and 15+ research papers. He was the convener for national workshop on Data Science and Big Data Analytics which was funded by DST, Govt. of India. He chaired sessions in various international/national conferences. He completed many online certification courses in Python, Software and Agile practices from Coursera offered by Google, University of Michigan, and University of Alberta. He has more than 16+ years of teaching and 9+ years of research experience in Engineering Institutions like VIT-AP University, GITAM University,
SVECW, etc. He is editorial member and reviewer of various peer-reviewed journals which are SCI/SCOPUS indexed like IET Software, JCS, IJECE, IJCS, BEEI, IJSER. He is a program committee member of many international conferences like ICCIDE2021, ACSTY2021, SOFE2021,
CSITA2021, SEAS2021, JSE2020 He is currently working on Imbalanced data and Data privacy in the application domains of Health Care, Smart cities, Financial, Agriculture and Social data analysis. He is an IEEE Member, life member in CSI, IAENG and AMIE.
\end{IEEEbiography}

\end{document}